\documentclass[11pt,a4paper]{article}
\usepackage[hyperref]{naaclhlt2019}
\usepackage{times}
\usepackage{latexsym}
\usepackage{CJKutf8}
\usepackage{booktabs}
\usepackage{multirow}
\usepackage{amsmath}
\usepackage{amssymb}
\usepackage{url}
\usepackage{graphicx}
\usepackage{helvet}  
\usepackage{courier}
\usepackage{threeparttable}
\usepackage{amsthm}
\theoremstyle{definition}

\def\tran{^\mathrm{\scriptscriptstyle T}}
\aclfinalcopy 

\title{Utilizing BERT for Aspect-Based Sentiment Analysis \\via Constructing Auxiliary Sentence}

\author{Chi Sun, Luyao Huang, Xipeng Qiu\thanks{{ }{ }Corresponding author.}\\
	Shanghai Key Laboratory of Intelligent Information Processing, Fudan University\\
	School of Computer Science, Fudan University\\
	825 Zhangheng Road, Shanghai, China\\
	{\tt \{sunc17,lyhuang18,xpqiu\}@fudan.edu.cn} \\}

\date{}

\begin{document}
	\maketitle
	\begin{abstract}
	Aspect-based sentiment analysis (ABSA), which aims to identify fine-grained opinion polarity towards a specific aspect, is a challenging subtask of sentiment analysis (SA). In this paper, we construct an auxiliary sentence from the aspect and convert ABSA to a sentence-pair classification task, such as question answering (QA) and natural language inference (NLI). We fine-tune the pre-trained model from BERT and achieve new state-of-the-art results on SentiHood and SemEval-2014 Task 4 datasets.\footnote{The source codes are available at \url{https://github.com/HSLCY/ABSA-BERT-pair}}
	\end{abstract}
	
	\section{Introduction}
	Sentiment analysis (SA) is an important task in natural language processing. It solves the computational processing of opinions, emotions, and subjectivity - sentiment is collected, analyzed and summarized. It has received much attention not only in academia but also in industry, providing real-time feedback through online reviews on websites such as Amazon, which can take advantage of customers' opinions on specific products or services. The underlying assumption of this task is that the entire text has an overall polarity.
	
	However, the users' comments may contain different aspects, such as: ``This book is a hardcover version, but the price is a bit high." The polarity in `appearance' is positive, and the polarity regarding `price' is negative. Aspect-based sentiment analysis (ABSA) \cite{jo2011aspect,S14-2004,pontiki2015semeval,pontiki2016semeval} aims to identify fine-grained polarity towards a specific aspect. This task allows users to evaluate aggregated sentiments for each aspect of a given product or service and gain a more granular understanding of their quality.
	
	Both SA and ABSA are sentence-level or document-level tasks, but one comment may refer to more than one object, and sentence-level tasks cannot handle sentences with multiple targets. Therefore, \citet{saeidi2016sentihood} introduce the task of targeted aspect-based sentiment analysis (TABSA), which aims to identify fine-grained opinion polarity towards a specific aspect associated with a given target. The task can be divided into two steps: (1) the first step is to determine the aspects associated with each target; (2) the second step is to resolve the polarity of aspects to a given target.
	
	The earliest work on (T)ABSA relied heavily on feature engineering \cite{wagner2014dcu,kiritchenko2014nrc}, and subsequent neural network-based methods \cite{nguyen2015phrasernn,wang2016attention,tang2015effective,tang2016aspect,wang2017tdparse} achieved higher accuracy. Recently, \citet{ma2018targeted} incorporate useful commonsense knowledge into a deep neural network to further enhance the result of the model. \citet{liu2018recurrent} optimize the memory network and apply it to their model to better capture linguistic structure.

	More recently, the pre-trained language models, such as ELMo \cite{peters2018deep}, OpenAI GPT \cite{radford2018improving}, and BERT \cite{devlin2018bert}, have shown their effectiveness to alleviate the effort of feature engineering. Especially, BERT has achieved excellent results in QA and NLI. However, there is not much improvement in (T)ABSA task with the direct use of the pre-trained BERT model (see Table \ref{table_sentihood}). We think this is due to the inappropriate use of the pre-trained BERT model.

	Since the input representation of BERT can represent both a single text sentence and a pair of text sentences, we can convert (T)ABSA into a sentence-pair classification task and fine-tune the pre-trained BERT.
		
	In this paper, we investigate several methods of constructing an auxiliary sentence and transform (T)ABSA into a sentence-pair classification task. We fine-tune the pre-trained model from BERT and achieve new state-of-the-art results on (T)ABSA task. We also conduct a comparative experiment to verify that the classification based on a sentence-pair is better than the single-sentence classification with fine-tuned BERT, which means that the improvement is not only from BERT but also from our method. In particular, our contribution is two-fold:
	
	1. We propose a new solution of (T)ABSA by converting it to a sentence-pair classification task.
	
	2. We fine-tune the pre-trained BERT model and achieve new state-of-the-art results on SentiHood and SemEval-2014 Task 4 datasets.
	
	\section{Methodology}
	In this section, we describe our method in detail.

	\subsection{Task description}

	\paragraph{TABSA}	In TABSA, a sentence $s$ usually consists of a series of words: $\{w_1,\cdots,w_m\}$, and some of the words $\{w_{i_1},\cdots,w_{i_k}\}$ are pre-identified targets $\{t_1,\cdots,t_k\}$, following \citet{saeidi2016sentihood}, we set the task as a 3-class classification problem: given the sentence $s$, a set of target entities $T$ and a fixed aspect set $A=\{general,price,transit\-location,safety\}$, predict the sentiment polarity $y\in\{positive, negative, none\}$ over the full set of the target-aspect pairs $\{(t, a): t \in T, a \in A\}$. As we can see in Table \ref{example}, the gold standard polarity of (LOCATION2, price) is negative, while the polarity of (LOCATION1, price) is none.

	\paragraph{ABSA} In ABSA, the target-aspect pairs $\{t,a\}$ become only aspects $a$. This setting is equivalent to learning subtasks 3 (Aspect Category Detection) and subtask 4 (Aspect Category Polarity) of SemEval-2014 Task 4\footnote{http://alt.qcri.org/semeval2014/task4/} at the same time.

	\begin{table}[ht]
		\centering
		\begin{tabular}{l c c}
			\toprule
            \textbf{Example:}\\
            \multicolumn{3}{p{\linewidth }}{\textcolor[rgb]{1.00,0.00,0.00}{LOCATION2} is central London so extremely expensive, \textcolor[rgb]{0.00,0.00,1.00}{LOCATION1} is often considered the coolest area of London.}\\
            \midrule
            \midrule
			Target & Aspect & Sentiment \\
			\cline{1-3}
			LOC1 & general & Positive \\
			LOC1 & price & None \\
			LOC1 & safety & None \\
			LOC1 & transit-location & None \\
			LOC2 & general & None \\
			LOC2 & price & Negative \\
			LOC2 & safety & None \\
			LOC2 & transit-location & Positive \\
			\bottomrule
		\end{tabular}
	\caption{\label{example} An example of SentiHood dataset.}
	\end{table}
	
    \subsection{Construction of the auxiliary sentence} \label{construct}
	For simplicity, we mainly describe our method with TABSA as an example.
	
	We consider the following four methods to convert the TABSA task into a sentence pair classification task:
	
	\begin{table}[ht]
		\centering
		\begin{tabular}{c c c}
			\toprule
			Methods & Output & Auxiliary Sentence  \\
			\cline{1-3}
			QA-M & S.P. & Question w/o S.P.\\			
			NLI-M & S.P. & Pseudo-sentence w/o S.P. \\
			QA-B & \{yes,no\} & Question w/ S.P. \\
			NLI-B & \{yes,no\} & Pseudo-sentence w/ S.P.\\
			\bottomrule
		\end{tabular}
		\caption {The construction methods. Due to limited space, we use the following abbreviations: \textit{S.P.} for \textit{sentiment polarity}, \textit{w/o} for \textit{without}, and \textit{w/} for \textit{with}. }
	\end{table}
	
	\paragraph{Sentences for QA-M}
	The sentence we want to generate from the target-aspect pair is a question, and the format needs to be the same. For example, for the set of a target-aspect pair (LOCATION1, safety), the sentence we generate is ``what do you think of the safety of location - 1 ?"
	
	\paragraph{Sentences for NLI-M}
	For the NLI task, the conditions we set when generating sentences are less strict, and the form is much simpler. The sentence created at this time is not a standard sentence, but a simple pseudo-sentence, with (LOCATION1, safety) pair as an example: the auxiliary sentence is: ``location - 1 -  safety".
	
	\paragraph{Sentences for QA-B}
	For QA-B, we add the label information and temporarily convert TABSA into a binary classification problem ($label\in\{yes, no\} $) to obtain the probability distribution. At this time, each target-aspect pair will generate three sequences such as ``the polarity of the aspect safety of location - 1 is positive", ``the polarity of the aspect safety of location - 1 is negative", ``the polarity of the aspect safety of location - 1 is none". We use the probability value of $yes$ as the matching score. For a target-aspect pair which generates three sequences ($positive, negative, none$), we take the class of the sequence with the highest matching score for the predicted category.
	
	\paragraph{Sentences for NLI-B}
	The difference between NLI-B and QA-B is that the auxiliary sentence changes from a question to a pseudo-sentence. The auxiliary sentences are: ``location - 1 - safety - positive", ``location - 1 - safety - negative", and ``location - 1 - safety - none".
	
	After we construct the auxiliary sentence, we can transform the TABSA task from a single sentence classification task to a sentence pair classification task. As shown in Table \ref{table_sentihood}, this is a necessary operation that can significantly improve the experimental results of the TABSA task.
	
	\subsection{Fine-tuning pre-trained BERT}
	BERT \cite{devlin2018bert} is a new language representation model, which uses bidirectional transformers to pre-train a large corpus, and fine-tunes the pre-trained model on other tasks. We fine-tune the pre-trained BERT model on TABSA task. Let's take a brief look at the input representation and the fine-tuning procedure.
	
	\subsubsection{Input representation} The input representation of the BERT can explicitly represent a pair of text sentences in a sequence of tokens. For a given token, its input representation is constructed by summing the corresponding token, segment, and position embeddings. For classification tasks, the first word of each sequence is a unique classification embedding ([CLS]).
	
	\subsubsection{Fine-tuning procedure} BERT fine-tuning is straightforward. To obtain a fixed-dimensional pooled representation of the input sequence, we use the final hidden state (i.e., the output of the transformer) of the first token as the input. We denote the vector as $C \in \mathbb{R}^H$. Then we add a classification layer whose parameter matrix is $W \in \mathbb{R}^{K \times H}$, where $K$ is the number of categories. Finally, the probability of each category $P$ is calculated by the softmax function $P=\mathbf{softmax}(CW\tran)$.
	
	\subsubsection{BERT-single and BERT-pair}
	\subparagraph{BERT-single for (T)ABSA} BERT for single sentence classification tasks. Suppose the number of target categories are $n_t$ and aspect categories are $n_a$. We consider TABSA as a combination of $n_t \cdot n_a$ target-aspect-related sentiment classification problems, first classifying each sentiment classification problem, and then summarizing the results obtained. For ABSA, We fine-tune pre-trained BERT model to train $n_a$ classifiers for all aspects and then summarize the results.
	
	\subparagraph{BERT-pair for (T)ABSA} BERT for sentence pair classification tasks. Based on the auxiliary sentence constructed in Section \ref{construct}, we use the sentence-pair classification approach to solve (T)ABSA. Corresponding to the four ways of constructing sentences, we name the models: BERT-pair-QA-M, BERT-pair-NLI-M, BERT-pair-QA-B, and BERT-pair-NLI-B.

	\section{Experiments}
		\begin{table*}[t!]
		\centering
		\begin{tabular}{l c c c c c c}
			\toprule
			\multirow{2}*{Model} & \multicolumn{3}{c}{Aspect} & & \multicolumn{2}{c}{Sentiment}\\
			\cline{2-4}
			\cline{6-7}
			~ & $Acc.$ & $F_{1}$ & AUC &  & $Acc.$ & AUC \\
			\midrule
			LR \cite{saeidi2016sentihood} & - & 39.3 & 92.4 & & 87.5 & 90.5 \\
			LSTM-Final \cite{saeidi2016sentihood} & - & 68.9 & 89.8 & & 82.0 & 85.4\\
			LSTM-Loc \cite{saeidi2016sentihood} & - & 69.3 & 89.7 & & 81.9 & 83.9\\
			LSTM+TA+SA \cite{ma2018targeted} & 66.4 & 76.7 & - & & 86.8 & -\\
			SenticLSTM \cite{ma2018targeted} & 67.4 & 78.2 & - & & 89.3 & -\\
			Dmu-Entnet \cite{liu2018recurrent} & 73.5 & 78.5 & 94.4 & & 91.0 & 94.8\\
			\midrule
			BERT-single & 73.7 & 81.0 & 96.4 & & 85.5 & 84.2 \\
			BERT-pair-QA-M & 79.4 & 86.4 & 97.0 & & \textbf{93.6} & 96.4 \\
			BERT-pair-NLI-M & 78.3 & 87.0 & \textbf{97.5} & & 92.1 & 96.5 \\
			BERT-pair-QA-B & 79.2 & \textbf{87.9} & 97.1 & & 93.3 & \textbf{97.0} \\
			BERT-pair-NLI-B & \textbf{79.8} & 87.5 & 96.6 & & 92.8 & 96.9 \\
			\bottomrule
		\end{tabular}
		\caption{\label{table_sentihood} Performance on SentiHood dataset. We boldface the score with the best performance across all models. We use the results reported in \citet{saeidi2016sentihood}, \citet{ma2018targeted} and \citet{liu2018recurrent}. ``-" means not reported.
		}
	\end{table*}
	\subsection{Datasets}
	We evaluate our method on the SentiHood \cite{saeidi2016sentihood} dataset\footnote{Dataset mirror: https://github.com/uclmr/jack/tree/master\\/data/sentihood}, which consists of 5,215 sentences, 3,862 of which contain a single target, and the remainder multiple targets.
	 Each sentence contains a list of target-aspect pairs $\{t,a\}$ with the sentiment polarity $y$. Ultimately, given a sentence $s$ and the target $t$ in the sentence, we need to:
	
	(1) detect the mention of an aspect $a$ for the target $t$;
	
	(2) determine the positive or negative sentiment polarity $y$ for detected target-aspect pairs.
	
	We also evaluate our method on SemEval-2014 Task 4 \cite{S14-2004} dataset\footnote{http://alt.qcri.org/semeval2014/task4/} for aspect-based sentiment analysis. The only difference from the SentiHood is that the target-aspect pairs $\{t,a\}$ become only aspects $a$. This setting allows us to jointly evaluate subtask 3 (Aspect Category Detection) and subtask 4 (Aspect Category Polarity).
	
	\subsection{Hyperparameters}
	We use the pre-trained uncased BERT-base model\footnote{https://storage.googleapis.com/bert\_models/2018\_10\_18/\\uncased\_L-12\_H-768\_A-12.zip} for fine-tuning. The number of Transformer blocks is 12, the hidden layer size is 768, the number of self-attention heads is 12, and the total number of parameters for the pre-trained model is 110M. When fine-tuning, we keep the dropout probability at 0.1, set the number of epochs to 4. The initial learning rate is 2e-5, and the batch size is 24.
	
	\subsection{Exp-I: TABSA}
	We compare our model with the following models:
	\begin{itemize}
		\item
		LR \cite{saeidi2016sentihood}: a logistic regression classifier with n-gram and pos-tag features.
		\item
		LSTM-Final \cite{saeidi2016sentihood}: a biLSTM model with the final state as a representation.
		\item
		LSTM-Loc \cite{saeidi2016sentihood}: a biLSTM model with the state associated with the target position as a representation.
		\item
		LSTM+TA+SA \cite{ma2018targeted}: a biLSTM model which introduces complex target-level and sentence-level attention mechanisms.
		\item
		SenticLSTM \cite{ma2018targeted}: an upgraded version of the LSTM+TA+SA model which introduces external information from SenticNet \cite{cambria2016senticnet}.
		\item
		Dmu-Entnet \cite{liu2018recurrent}: a bi-directional EntNet \cite{henaff2016tracking} with external “memory chains” with a delayed memory update mechanism to track entities.
		
	\end{itemize}
	During the evaluation of SentiHood, following \citet{saeidi2016sentihood}, we only consider the four most frequently seen aspects (general, price, transit-location, safety). When evaluating the aspect detection, following \citet{ma2018targeted}, we use strict accuracy and Macro-F1, and we also report AUC.  In sentiment classification, we use accuracy and macro-average AUC as the evaluation indices.
	
	\subsubsection{Results}
	Results on SentiHood are presented in Table \ref{table_sentihood}. The results of the BERT-single model on aspect detection are better than Dmu-Entnet, but the accuracy of sentiment classification is much lower than that of both SenticLstm and Dmu-Entnet, with a difference of 3.8 and 5.5 respectively.
	
	However, BERT-pair outperforms other models on aspect detection and sentiment analysis by a substantial margin, obtaining 9.4 macro-average F1 and 2.6 accuracies improvement over Dmu-Entnet. Overall, the performance of the four BERT-pair models is close. It is worth noting that BERT-pair-NLI models perform relatively better on aspect detection, while BERT-pair-QA models perform better on sentiment classification. Also, the BERT-pair-QA-B and BERT-pair-NLI-B models can achieve better AUC values on sentiment classification than the other models.

	\subsection{Exp-II: ABSA}

	The benchmarks for SemEval-2014 Task 4 are the two best performing systems in \citet{S14-2004} and ATAE-LSTM \cite{wang2016attention}. When evaluating SemEval-2014 Task 4 subtask 3 and subtask 4, following \citet{S14-2004}, we use Micro-F1 and accuracy respectively.
	\subsubsection{Results}
	Results on SemEval-2014 are presented in Table \ref{table_semeval_2014_4_3} and Table \ref{table_semeval_2014_4_4}. We find that BERT-single has achieved better results on these two subtasks, and BERT-pair has achieved further improvements over BERT-single. The BERT-pair-NLI-B model achieves the best performance for aspect category detection. For aspect category polarity, BERT-pair-QA-B performs best on all 4-way, 3-way, and binary settings.
	
	\begin{table}[t!]
		\centering
		\begin{tabular}{l c c c}
			\toprule
			Models & P & R & F1  \\
			\midrule		
			XRCE & 83.23 & 81.37 & 82.29 \\
			NRC-Canada & 91.04 & 86.24 & 88.58 \\
			\midrule
			BERT-single & 92.78 & 89.07 & 90.89 \\
			BERT-pair-QA-M & 92.87 & 90.24 & 91.54 \\
			BERT-pair-NLI-M & 93.15 & 90.24 & 91.67 \\
			BERT-pair-QA-B & 93.04 & 89.95 & 91.47 \\
			BERT-pair-NLI-B & 93.57 & 90.83 & \textbf{92.18} \\
			\bottomrule
		\end{tabular}
		\caption{\label{table_semeval_2014_4_3} Test set results for Semeval-2014 task 4 Subtask 3: Aspect Category Detection. We use the results reported in XRCE \cite{brun2014xrce} and NRC-Canada \cite{kiritchenko2014nrc}.
		}
	\end{table}

	\begin{table}[t!]
		\centering
		\begin{tabular}{l c c c}
			\toprule
			Models & 4-way & 3-way & Binary  \\
			\midrule
			XRCE & 78.1 & - & - \\			
			NRC-Canada & 82.9 & - & - \\
			LSTM & - & 82.0 & 88.3 \\
			ATAE-LSTM & - & 84.0 & 89.9 \\
			\midrule
			BERT-single & 83.7 & 86.9 & 93.3 \\
			BERT-pair-QA-M & 85.2 & 89.3 & 95.4 \\
			BERT-pair-NLI-M & 85.1 & 88.7 & 94.4 \\
			BERT-pair-QA-B & \textbf{85.9} & \textbf{89.9} & \textbf{95.6} \\
			BERT-pair-NLI-B & 84.6 & 88.7 & 95.1 \\
			\bottomrule
		\end{tabular}
		\caption{\label{table_semeval_2014_4_4} Test set accuracy (\%) for Semeval-2014 task 4 Subtask 4: Aspect Category Polarity. We use the results reported in XRCE \cite{brun2014xrce}, NRC-Canada \cite{kiritchenko2014nrc} and ATAE-LSTM \cite{wang2016attention}. ``-" means not reported.
		}
	\end{table}

	\section{Discussion}
	
	Why is the experimental result of the BERT-pair model so much better? On the one hand, we convert the target and aspect information into an auxiliary sentence, which is equivalent to exponentially expanding the corpus. A sentence $s_i$ in the original data set will be expanded into $(s_i, t_1, a_1),\cdots,(s_i, t_1, a_{n_a}),\cdots,(s_i, t_{n_t}, a_{n_a})$ in the sentence pair classification task. On the other hand, it can be seen from the amazing improvement of the BERT model on the QA and NLI tasks \cite{devlin2018bert} that the BERT model has an advantage in dealing with sentence pair classification tasks. This advantage comes from both unsupervised masked language model and next sentence prediction tasks.
	
	TABSA is more complicated than SA due to additional target and aspect information. Directly fine-tuning the pre-trained BERT on TABSA does not achieve performance growth. However, when we separate the target and the aspect to form an auxiliary sentence and transform the TABSA into a sentence pair classification task, the scenario is similar to QA and NLI, and then the advantage of the pre-trained BERT model can be fully utilized. Our approach is not limited to TABSA, and this construction method can be used for other similar tasks. For ABSA, we can use the same approach to construct the auxiliary sentence with only aspects.
	
	In BERT-pair models, BERT-pair-QA-B and BERT-pair-NLI-B achieve better AUC values on sentiment classification, probably because of the modeling of label information.
	
	\section{Conclusion}
	In this paper, we constructed an auxiliary sentence to transform (T)ABSA from a single sentence classification task to a sentence pair classification task. We fine-tuned the pre-trained BERT model on the sentence pair classification task and obtained the new state-of-the-art results. We compared the experimental results of single sentence classification and sentence pair classification based on BERT fine-tuning, analyzed the advantages of sentence pair classification, and verified the validity of our conversion method. In the future, we will apply this conversion method to other similar tasks.
	
	\section*{Acknowledgments}
	We would like to thank the anonymous reviewers for their valuable comments. The research work is supported by Shanghai Municipal Science and Technology Commission (No. 16JC1420401 and 17JC1404100),
	National Key Research and Development Program of China (No. 2017YFB1002104),
	and National Natural Science Foundation of China (No. 61672162 and 61751201).
	
	\bibliography{sunnysc_target}
	\bibliographystyle{acl_natbib}
\end{document}